\pdfoutput=1

\documentclass[11pt]{article}

\usepackage[]{ACL2023}

\usepackage{times}
\usepackage{latexsym}
\usepackage{booktabs}
\usepackage{float}
\usepackage{amssymb}
\usepackage{makecell}
\usepackage{graphicx}
\usepackage{multirow}
\usepackage{subcaption}
\usepackage{array}
\usepackage{arydshln}

\usepackage[T1]{fontenc}

\usepackage[utf8]{inputenc}

\usepackage{microtype}

\usepackage{inconsolata}

\title{Gemini in Reasoning: Unveiling Commonsense in \\ Multimodal Large Language Models}

\author{Yuqing Wang \\
  Stanford University \\
  \texttt{ywang216@stanford.edu} \\\And
  Yun Zhao \\
  Meta Platforms, Inc. \\
  \texttt{yunzhao20@meta.com} \\}

\begin{document}
\maketitle
\begin{abstract}
The burgeoning interest in Multimodal Large Language Models (MLLMs), such as OpenAI's GPT-4V(ision), has significantly impacted both academic and industrial realms. These models enhance Large Language Models (LLMs) with advanced visual understanding capabilities, facilitating their application in a variety of multimodal tasks. Recently, Google introduced Gemini, a cutting-edge MLLM designed specifically for multimodal integration. Despite its advancements, preliminary benchmarks indicate that Gemini lags behind GPT models in commonsense reasoning tasks. However, this assessment, based on a limited dataset (i.e., HellaSWAG), does not fully capture Gemini's authentic commonsense reasoning potential. To address this gap, our study undertakes a thorough evaluation of Gemini's performance in complex reasoning tasks that necessitate the integration of commonsense knowledge across modalities. We carry out a comprehensive analysis of 12 commonsense reasoning datasets, ranging from general to domain-specific tasks. This includes 11 datasets focused solely on language, as well as one that incorporates multimodal elements. Our experiments across four LLMs and two MLLMs demonstrate Gemini's competitive commonsense reasoning capabilities. Additionally, we identify common challenges faced by current LLMs and MLLMs in addressing commonsense problems, underscoring the need for further advancements in enhancing the commonsense reasoning abilities of these models. Our data and results are available at: \url{https://github.com/EternityYW/Gemini-Commonsense-Evaluation/}.
\end{abstract}

\section{Introduction}
Commonsense reasoning, integral to human cognition, plays a crucial role in navigating the intricacies of everyday life. Consider a scenario where someone decides what to wear based on the weather. This decision extends beyond the mere selection of attire; it involves understanding weather patterns, the suitability of clothing for different temperatures, and the social context of the occasion. It's about synthesizing diverse pieces of knowledge: a forecast predicting rain, the practical necessity for a raincoat, and the societal expectation of dressing appropriately for an event. This reasoning goes beyond simply processing information; it entails integrating varied pieces of knowledge that humans often take for granted. A major challenge in Natural Language Processing (NLP) research is the ambiguity and under-specification of human language. Individuals rely heavily on their commonsense knowledge and reasoning abilities to decipher these ambiguities and infer missing information. Commonsense reasoning has consistently posed unique challenges in NLP research~\citep{li2021language, bian2023chatgpt}, encompassing spatial, physical, social, temporal, and psychological aspects, along with an understanding of social norms, beliefs, values, and the nuances of predicting and interpreting human behavior~\citep{liu2004conceptnet}. Models often lack this innate commonsense, hindering their ability to contextualize data coherently, in stark contrast to the human capacity for effortlessly understanding everyday situations~\citep{shwartz2020neural, bhargava2022commonsense}.

Recent advances in Large Language Models (LLMs) have sparked unprecedented enthusiasm in the NLP community and beyond, significantly enhancing a wide array of applications~\citep{min2021recent, zhao2023survey, wang2023large, kasneci2023chatgpt, he2023large}. Building on these achievements, Multimodal Large Language Models (MLLMs) have emerged as a pivotal focus in the next wave of AI~\citep{wu2023multimodal}, speculated to advance towards Artificial General Intelligence (AGI), which aims to develop AI systems smarter than humans and beneficial for all of humanity~\citep{rayhan2023artificial}. The rise of MLLMs, particularly OpenAI's GPT-4V(ision)~\citep{yang2023dawn} and Google's Gemini~\citep{team2023gemini}, marks significant progress in this area. Among these developments, Gemini emerges as a formidable challenger to the state-of-the-art MLLM, GPT-4V, specially engineered for multimodal integration. Its release has ignited constructive discussions about the current level of AGI achievement. In widely used academic benchmarks, Gemini has attained new state-of-the-art status in the majority of tasks. However, preliminary evaluations of Gemini, especially when compared to models like the GPT series, have indicated potential shortcomings in its commonsense reasoning capabilities, a fundamental aspect of human cognition. Yet, it's important to consider that basing the assessment of Gemini's commonsense reasoning abilities solely on the HellaSWAG dataset~\citep{zellers2019hellaswag} may not comprehensively reflect Gemini's full scope in this critical domain. 

To address the gap in the comprehensive evaluation of Gemini's real-world performance in commonsense reasoning tasks, our study conducts extensive experiments across 12 commonsense reasoning datasets, covering a broad spectrum of domains such as general, physical, social, and temporal reasoning. We experiment with four popular LLMs for the language dataset evaluation, including Llama2-70b~\citep{touvron2023llama}, Gemini Pro~\citep{team2023gemini}, GPT-3.5 Turbo, and GPT-4 Turbo~\citep{openai2023gpt4}. For the multimodal dataset, we assess both Gemini Pro Vision and GPT-4V. Our key findings are summarized as follows: (1) Overall, Gemini Pro's performance is comparable to that of GPT-3.5 Turbo, demonstrating marginally better average results across 11 language datasets (1.4\% higher accuracy), though it lags behind GPT-4 Turbo by an average of 8.2\% in accuracy. Moreover, Gemini Pro Vision exhibits lower performance than GPT-4V on the multimodal dataset, except for temporal-related questions. (2) Approximately 65.8\% of Gemini Pro's reasoning processes are evaluated as logically sound and contextually relevant, indicating its potential for effective application in various domains. (3) Gemini Pro encounters significant challenges in temporal and social commonsense reasoning, indicating key areas for further development. (4) Our manual error analysis reveals that Gemini Pro often misunderstands provided contextual information, accounting for 30.2\% of its total errors. Furthermore, Gemini Pro Vision struggles particularly with identifying emotional stimuli in images, especially those involving human entities, which constitutes 32.6\% of its total errors.

In summary, our contributions are threefold:
\begin{enumerate}
\item[(1)] We provide the first thorough evaluation of Gemini Pro's efficacy in commonsense reasoning tasks, employing 12 diverse datasets that span both language-based and multimodal scenarios.
\item[(2)] Our study reveals that Gemini Pro exhibits performance comparable to GPT-3.5 Turbo in language-only commonsense reasoning tasks, demonstrating logical and contextual reasoning processes. However, it lags behind GPT-4 Turbo in accuracy and encounters challenges in temporal and social reasoning, as well as in emotion recognition in images.

\item[(3)] Our findings lay a valuable foundation for future research in the field of commonsense reasoning within LLMs and MLLMs, highlighting the necessity to enhance specialized domains in these models and the nuanced recognition of mental states and emotions in multimodal contexts.
\end{enumerate}

\section{Commonsense Overview}
Commonsense reasoning, a fundamental aspect of human intelligence, facilitates an intuitive understanding and interpretation of the world through basic and often implicit knowledge and beliefs. For instance, it involves understanding that a person carrying an umbrella on a cloudy day likely anticipates rain, or inferring that a closed door in a library signifies a need for quiet. In MLLMs, commonsense reasoning plays a vital role, enabling these models to interact with and interpret human language and visual cues in a manner that mirrors human understanding. In our study, we explore a variety of commonsense reasoning tasks. Definitions for each domain are provided as follows.

\noindent \textbf{General Commonsense.} This domain entails an understanding of basic, everyday knowledge about the world, such as recognizing that birds typically fly and fish live in water.

\noindent \textbf{Contextual Commonsense.} This domain involves interpreting information within specific contexts, such as understanding that a person wearing a coat and shivering is likely cold.

\noindent \textbf{Abductive Commonsense.} This domain is about formulating the most plausible explanations for observations, such as inferring that wet streets are likely due to recent rain.

\noindent \textbf{Event Commonsense.} This domain focuses on understanding sequences of events and the causal relationships between them, such as predicting that eating spoiled food can lead to feeling sick.

\noindent \textbf{Temporal Commonsense.} This domain involves understanding time-related concepts, such as the fact that breakfast is typically eaten in the morning.

\noindent \textbf{Numerical Commonsense.} This domain is about understanding numbers in everyday contexts, such as knowing that a cube has six faces.

\noindent \textbf{Physical Commonsense.} This domain concerns understanding the physical world, such as knowing that a glass will break if dropped on a hard floor.

\noindent \textbf{Science Commonsense.} This domain involves the application of scientific principles in daily life, such as understanding that water boils at a higher temperature at sea level than in the mountains.

\noindent \textbf{Riddle Commonsense.} This domain challenges creative thinking through riddles, such as deciphering a riddle where the answer is “a shadow”, requiring lateral thinking to associate intangible concepts with physical entities.

\noindent \textbf{Social Commonsense.} This domain involves understanding social interactions, such as recognizing that a person is likely upset if he/she is crying.

\noindent \textbf{Moral Commonsense.} This domain deals with evaluating actions based on moral and ethical standards, such as understanding that stealing is generally considered wrong.

\noindent \textbf{Visual Commonsense.} This domain involves interpreting and understanding visual information in the context of the physical and social world, such as deducing that a person in a photo is likely running a race if they are wearing a number bib and surrounded by other runners.

\section{Experimental Setup}

\subsection{Datasets}~\label{data_section}
We experiment with 12 datasets related to different types of commonsense reasoning, which include 11 language-based datasets and one multimodal dataset. The language-based datasets encompass three main categories of commonsense reasoning problems: \textbf{General and Contextual Reasoning:} (1) CommonsenseQA~\citep{talmor2019commonsenseqa}, focusing on general commonsense knowledge; (2) Cosmos QA~\citep{huang2019cosmos}, emphasizing contextual understanding narratives, (3) $\alpha$NLI~\citep{bhagavatula2019abductive}, introducing abductive reasoning, which involves inferring the most plausible explanation; and
(4) HellaSWAG, centering around reasoning with contextual event sequences. \textbf{Specialized and Knowledge Reasoning:} (1) TRAM~\citep{wang2023tram}, testing reasoning about time; (2) NumerSense~\citep{lin2020birds}, focusing on numerical understanding; (3) PIQA~\citep{bisk2020piqa}, assessing physical interaction knowledge; (4) QASC~\citep{khot2020qasc}, dealing with science-related reasoning; and (5) RiddleSense~\citep{lin2021riddlesense}, challenging creative thinking through riddles. \textbf{Social and Ethical Reasoning:} (1) Social IQa~\citep{sap2019social}, testing the understanding of social interactions; and (2) ETHICS~\citep{hendrycks2020aligning}, evaluating moral and ethical reasoning. For the multimodal dataset (vision and language), we select VCR~\citep{zellers2019recognition}, a large-scale dataset for cognition-level visual understanding. For datasets like TRAM and ETHICS, which include multiple tasks, we extract the commonsense reasoning part for experiments. We employ accuracy as the performance metric for all datasets. Table~\ref{tab: dataset} provides an overview of the datasets, as well as example questions.

\subsection{Models}
We consider four popular LLMs for language-based dataset evaluation, including the open-source model Llama-2-70b-chat~\citep{touvron2023llama} as well as the closed-source models Gemini Pro~\citep{team2023gemini}, GPT-3.5 Turbo, and GPT-4 Turbo~\citep{openai2023gpt4}. Each of these models is accessed using its corresponding API key. Specifically, we query Gemini through Google Vertex AI, the GPT models through the OpenAI API, and Llama2 through DeepInfra. For the multimodal dataset, we consider GPT-4V (gpt-4-vision-preview in API) and Gemini Pro Vision (gemini-pro-vision in API) in our experiments. Given the constraints of API costs and rate limitations, we randomly select 200 examples from the validation set for each language-based dataset following~\citep{wang2023tram} and 50 examples from the validation set for the VCR dataset following~\citep{liu2023evaluation}. For all evaluations, we employ greedy decoding (i.e., temperature = 0) during model response generation. Notably, there are instances where the models decline to respond to certain queries, particularly those involving potentially illegal or unethical content. Sometimes, models provide answers that are outside the scope of the options. In these cases, we categorize these unanswered questions as incorrect.

\subsection{Prompts}
In the evaluation of language-based datasets, we employ two prompting settings: (1) zero-shot standard prompting (SP)~\citep{kojima2022large}, which aims to gauge the models' inherent commonsense capabilities in linguistic contexts, and (2) few-shot chain-of-thought (CoT) prompting~\citep{wei2022chain}, implemented to observe potential enhancements in the models' performance. For the multimodal dataset, we utilize zero-shot standard prompting to assess the authentic end-to-end visual commonsense reasoning abilities of MLLMs.

\begin{table*}[htbp]
\centering 
\caption{Overview of commonsense datasets used in our experiments. “K-Way MC” signifies a multiple-choice response format with K options. Bold text in the “Example Questions” column represents the correct answers.}
\resizebox{\linewidth}{!}{%
\renewcommand{\arraystretch}{1.25}
\begin{tabular}{cccc}
    \toprule
    \textbf{Dataset} & \textbf{Domain}  & \textbf{Answer Type} & \textbf{Example Questions} \\ 
    \midrule
    \multicolumn{4}{c}{General and Contextual Reasoning} \\
    \midrule
    CommonsenseQA & general & 5-Way MC & \makecell[l]{Where is a doormat likely to be in front of? \\ (A). facade; \textbf{(B). front door}; (C). doorway; (D). entrance porch; (E). hallway.}\\ \hdashline
    Cosmos QA & contextual & 4-Way MC & \makecell[l]{Given the context “It wasn't time for my book to be released... I have received \\ about five rejection letters.” What may be the reason for your book getting rejected? \\ (A). None of the above choices; (B). I never...; (C). I felt...; \textbf{(D). It wasn't finished.}} \\ \hdashline
    $\alpha$NLI & abductive & 2-Way MC & \makecell[l]{Given the beginning of the story: Four Outlaws camped in Blood Gulch, \\and the end of the story: He arrested them, what is the more plausible hypothesis: \\ (A). They found where the sheriff was; \textbf{(B). The sheriff found where they were.}}\\ \hdashline
    HellaSWAG & event & 4-Way MC & \makecell[l]{Given the context “A boy in an orange shirt is playing a video game. the scene” \\ and the activity label “Washing face”, which of the following endings is the most \\ appropriate continuation of the scenario? (A). changes to safety features; \\ \textbf{(B). changes to the game itself}; (C). switches to show...; (D). cuts to the boys...} \\
    \midrule
    \multicolumn{4}{c}{Specialized and Knowledge Reasoning} \\
    \midrule
    TRAM & temporal & 3-Way MC & \makecell[l]{Then the green ball told the orange ball that blue ball was stupid. How long was the \\ green ball talking to the orange ball? (A). 5 weeks; (B). 24 hours; \textbf{(C). 15 seconds}.} \\ \hdashline
    NumerSense & numerical & Number & \makecell[l]{Complete the sentence by filling in <mask> with the most appropriate number. \\ A classical guitar has <mask> strings. $\rightarrow$ \textbf{six}}\\ \hdashline
    PIQA & physical & 2-Way MC & \makecell[l]{To reach the physical goal: trees, choose the more sensible solution: \\ (A). provide homes for people;  \textbf{(B). provide homes for animals.}} \\ \hdashline
    QASC & science & 8-Way MC & \makecell[l]{Crabs scavanage and uses dead snail shells for what? (A). RNA; \textbf{(B). homes}; (C). making \\ holes; (D). damage; (E). a hosta; (F). Protein; (G). matter; (H). building a nest.}\\ \hdashline
    RiddleSense & riddle & 5-Way MC & \makecell[l]{Something very helpful if you want to go gently down a stream. (A). raft; (B). roll \\ down hill; \textbf{(C). rowboat}; (D). water; (E). roll over.}\\
    \midrule
    \multicolumn{4}{c}{Social and Ethical Reasoning} \\
    \midrule
     Social IQa & social & 3-Way MC & \makecell[l]{Kai observed and improved Jan's work. What will Kai want to do next? \\ (A). be effective; \textbf{(B). get praise}; (C). get incredible.} \\ \hdashline
     ETHICS & moral & 2-way MC & \makecell[l]{After I got the flu virus I called my friends to tell them that I had the virus. \\ Is this acceptable? \textbf{(A). acceptable}; (B). unacceptable.}\\
     \midrule
     \multicolumn{4}{c}{Visual Understanding and Reasoning} \\
     \midrule
     VCR & visual & 4-Way MC &     \begin{minipage}{0.42\textwidth}
  \includegraphics[width=\linewidth, height=50mm]{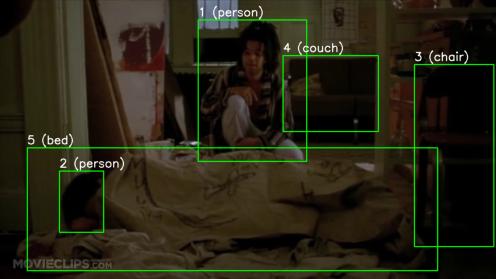}
\end{minipage}\hfill
\begin{minipage}{0.4\textwidth}
  1. What is wrong with Person 2? (A). He is not happy with what is being said to him over the telephone; \textbf{(B). He is feeling depressed}; (C). He is high on pot; (D). Someone has pushed him and he's falling. \\
2. Given the question: What is wrong with Person 2?, and the answer to the question: He is feeling depressed, what is the rationale behind this answer? \textbf{(A). Person 1 is talking to him probably trying to cheer him up}; (B). He looks sad and is drinking; (C). He is walking with his head down; (D). He is slumped down on bed and his eyes are closed.
    \end{minipage}\\
    \bottomrule
\end{tabular}}
\footnotesize 
\label{tab: dataset}
\end{table*}

\section{Results}

\subsection{Overall Performance Comparison}
Table~\ref{tab: overall_performance_language} demonstrates the accuracy comparison of four LLMs under zero-shot SP and few-shot CoT settings on 11 language-based commonsense reasoning datasets. There are several key takeaways. First, from the model perspective, GPT-4 Turbo outperforms the other models across the majority of datasets on average. Under the zero-shot learning paradigm, it surpasses Gemini Pro, the second-best performing model, by 7.3\%, and shows an even greater lead of 9.0\% under the few-shot learning paradigm. Gemini Pro exhibits marginally higher average accuracy than GPT-3.5 Turbo, with an increase of 1.3\% under zero-shot SP and 1.5\% in the few-shot CoT scenario. It also demonstrates substantially better performance than Llama-2-70b. Regarding prompting methods, the CoT approach consistently enhances performance across all datasets, with pronounced gains observed in datasets such as CommonsenseQA, TRAM, and Social IQa. Lastly, from a dataset standpoint, it is apparent that while these models exhibit commendable performance across a broad spectrum of commonsense domains, they encounter challenges in specific areas, particularly those involving temporal (TRAM) and social (Social IQa) dimensions of commonsense reasoning.

For the multimodal VCR dataset, we report the performance of GPT-4V and Gemini Pro Vision in Table~\ref{tab: overall_performance_MM}. The VCR consists of three subtasks: (1) Q $\rightarrow$ A, which involves generating an answer to a question based on the visual context; (2) QA $\rightarrow$ R, which requires the model to produce a rationale for a given answer; and (3) Q $\rightarrow$ AR, which challenges the model to both answer the question and justify the response with appropriate rationales. In all subtasks, GPT-4V demonstrates superior performance compared to Gemini Pro Vision, indicating a more robust capacity for integrating visual and textual information to provide coherent responses. In Q $\rightarrow$ AR, the relatively lower performance of both models, compared to the other two subtasks, suggests that there is considerable room for improvement in understanding the interplay between visual cues and commonsense reasoning.

\begin{table*}[h]
\centering 
\caption{Performance comparison of four LLMs across 11 language-based commonsense reasoning datasets. For the k-shot CoT setting, k is set to 5 for the majority of datasets, except HellaSWAG (k=10) and PIQA (k=1). The best results for the k-shot setting are boldfaced, and for the 0-shot setting, underlined. GPT-4 Turbo outperforms other models across the majority of datasets under both settings by a large margin. Gemini Pro and GPT-3.5 Turbo exhibit comparably matched performance overall, with Gemini Pro demonstrating marginally superior commonsense reasoning capabilities compared to GPT-3.5 Turbo on average.}
\resizebox{\linewidth}{!}{%
\renewcommand{\arraystretch}{1.15}
\begin{tabular}{cccccccccc}
 \toprule
  \multirow{3}{*}{\textbf{Dataset}} &  & &  &\textbf{Method} \\ \cmidrule(lr){2-10} 
   & Llama-2-70b & Llama-2-70b &  Gemini Pro & Gemini Pro & GPT-3.5 Turbo & GPT-3.5 Turbo & GPT-4 Turbo & GPT-4 Turbo\\ &  (0-shot, SP) & (k-shot, CoT) & (0-shot, SP) & (k-shot, CoT) & (0-shot, SP) & (k-shot, CoT) & (0-shot, SP) & (k-shot, CoT) \\
   \midrule
   CommonsenseQA & 72.0 & 76.5 & 76.5 & 79.0 & 73.0 & 76.0 & \underline{78.0} & \textbf{80.0}\\
   Cosmos QA & 77.0 & 81.0 & 81.5 & 84.5 & 75.0 & 78.5 & \underline{86.5} & \textbf{88.0} \\
   $\alpha$NLI & 77.5 & 80.5 & 79.5 & 81.5 & 75.5 & 78.0 & \underline{87.0} & \textbf{88.0} \\
   HellaSWAG & 73.0 & 77.0 & 76.0 & 78.5 & 78.0 & 80.0 & \underline{94.0} & \textbf{95.0}\\
   TRAM & 66.0 & 70.0 & 73.5 & 76.0 & 68.5 & 72.0 & \underline{79.5} & \textbf{82.0} \\
   NumerSense & 74.0 & 75.5 & 80.0 & 82.0 & 81.5 & 82.5 & \underline{85.0} & \textbf{86.0}\\
   PIQA & 74.0 & 78.5 & 89.0 & 90.5 & 87.0 & 89.5 & \underline{94.5} & \textbf{95.5}\\
   QASC & 78.0 & 82.0 & 80.0 & 82.5 & 83.0 & 85.0 & \underline{91.5} & \textbf{92.5} \\
   RiddleSense & 62.5 & 66.0 & 75.0 & 82.5 & 71.5 & 75.0 & \underline{94.0} & \textbf{95.0}\\
   Social IQa & 71.0 & 77.5 & 73.0 & 78.5 & 73.0 & 78.0 & \underline{82.0} & \textbf{84.5}\\
   ETHICS & 88.0 & 89.5 & 87.0 & 87.5 & 94.0 & 95.0 & \underline{97.0} & \textbf{98.0}\\
   \midrule
   Average & 73.9 & 77.6 & 79.2 & 82.1 & 78.2 & 80.9 & \underline{88.1} & \textbf{89.5} \\
   
   \bottomrule
\end{tabular}}
\label{tab: overall_performance_language}
\end{table*}

\begin{table}[h]
\centering 
\caption{Performance comparison between GPT-4V and Gemini Pro Vision on the VCR dataset. “Q $\rightarrow$ A” evaluates question-answering accuracy, “QA $\rightarrow$ R” assesses answer justification, and “Q $\rightarrow$ AR” measures the performance of both correctly answering questions and selecting rationales. GPT-4V outperforms Gemini Pro Vision across all subtasks.}
\resizebox{\linewidth}{!}{%
\begin{tabular}{cccc}
 \toprule
    \textbf{Method} & \textbf{Q $\rightarrow$ A} & \textbf{QA $\rightarrow$ R} & \textbf{Q $\rightarrow$ AR}\\ \midrule 
GPT-4V & 80.0 & 72.0 & 56.0 \\
Gemini Pro Vision & 74.0 & 70.0 & 48.0 \\
   \bottomrule
\end{tabular}}
\label{tab: overall_performance_MM}
\end{table}

\subsection{Effects of Commonsense Domain}
\begin{figure*}[h]
\centering
\includegraphics[width=\textwidth]{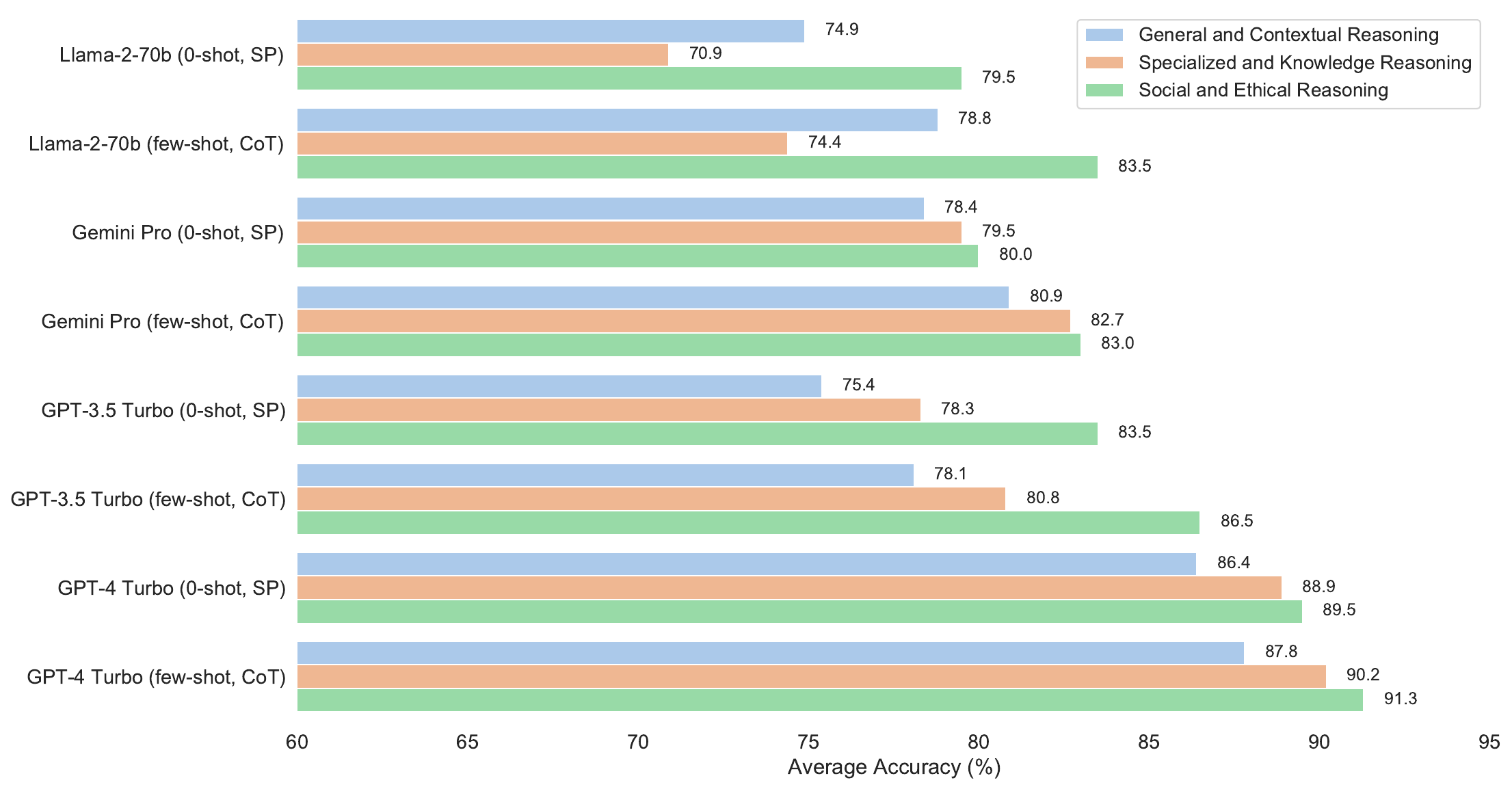}  
\caption{Average model performance across three major commonsense reasoning categories over 11 language-based datasets, including General and Contextual Reasoning (CommonsenseQA, Cosmos QA, $\alpha$NLI, HellaSWAG), Specialized and Knowledge Reasoning (TRAM, NumerSense, PIQA, QASC, RiddleSense), and Social and Ethical Reasoning (Social IQa, ETHICS). GPT-4 Turbo consistently exhibits superior performance in all commonsense reasoning categories. Gemini Pro marginally surpasses GPT-3.5 Turbo in the first two categories, except for Social and Ethical Reasoning.}
\label{fig: avg_perf_language_category}
\end{figure*}

Referring to Section~\ref{data_section}, we have categorized 11 language-based datasets into three groups and presented the performance for each setting within each group in Figure~\ref{fig: avg_perf_language_category}. Our findings indicate that GPT-4 Turbo consistently leads in performance across all categories. The Llama-2-70b model demonstrates marginally lower accuracy in comparison to the other models. Gemini Pro and GPT-3.5 Turbo display comparable performances; however, Gemini Pro slightly outperforms GPT-3.5 Turbo in two of the three categories. Notably, its performance dip in the Social and Ethical Reasoning group may stem from its tendency to refuse to answer questions that could potentially involve unethical content, which we have counted as incorrect in our evaluation. Based on our experiments, among the 200 samples, Gemini Pro refuses to answer 3.0\% of the problems (6 in total) in the Social IQa dataset and 6.5\% of the problems (13 in total) in the ETHICS dataset. Overall, all models exhibit robust capabilities in handling Social and Ethical Reasoning datasets, suggesting a relatively advanced grasp of moral and social norms. However, there is a notable disparity in their performance on General and Contextual Reasoning tasks, indicating a potential gap in their understanding of broader commonsense principles and their application in varied contexts. The Specialized and Knowledge Reasoning category, particularly in the realms of temporal and riddle-based challenges, highlights specific deficiencies in the models' abilities to process complex temporal sequences and to engage in the abstract and creative thought required to decipher riddles. 

\begin{figure*}[h]
\centering
\includegraphics[width=0.9\textwidth]{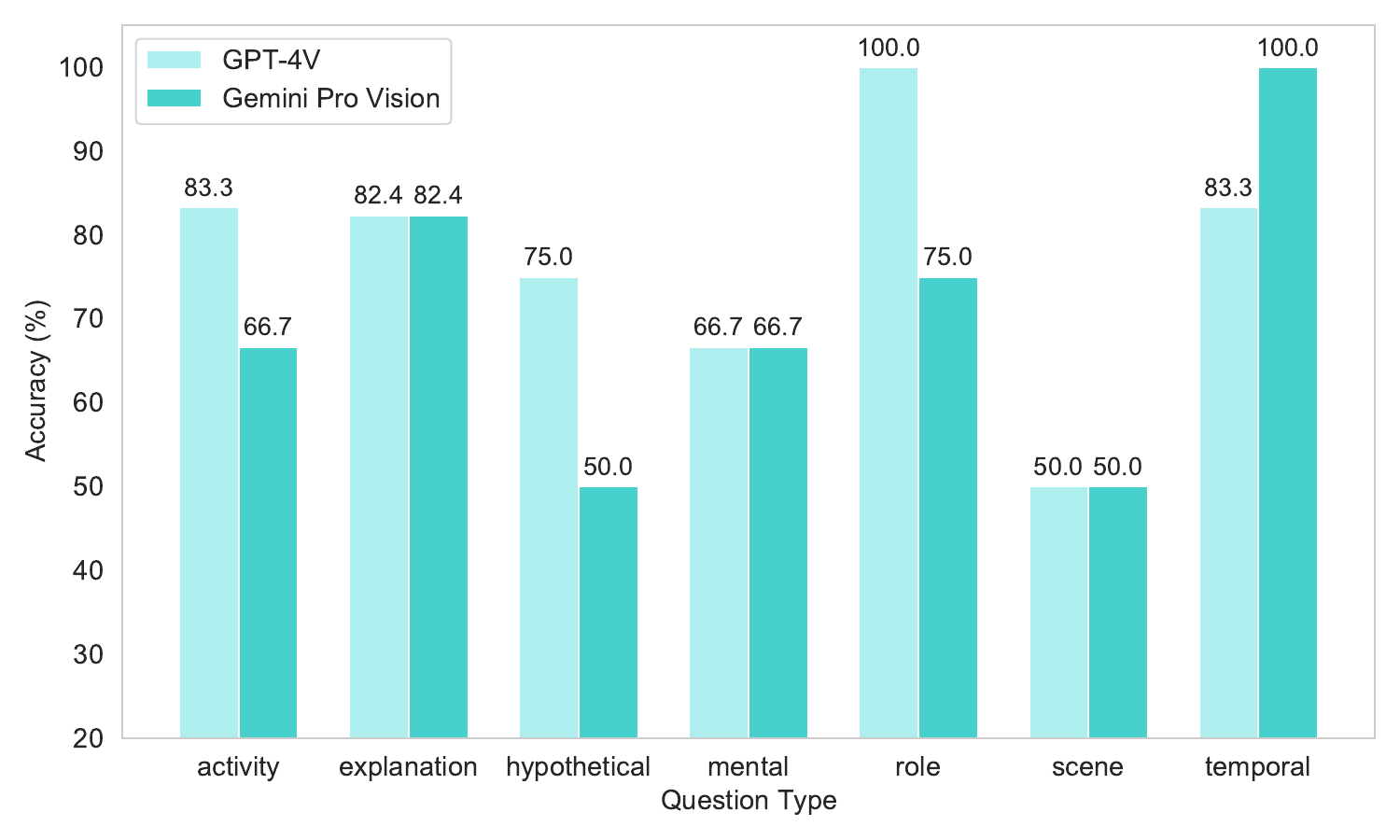}  
\caption{Performance comparison between GPT-4V and Gemini Pro Vision on the VCR dataset, categorized by question type, with a focus on the “Q $\rightarrow$ A” subtask. Within our sample of 50 questions, the distribution across each type is as follows: activity (12), explanation (16), hypothetical (3), mental (4), role (5), scene (4), and temporal (6). GPT-4V matches or surpasses Gemini Pro Vision in performance across these question types, with the exception of the temporal category.}
\label{fig: avg_perf_mm_category}
\end{figure*}

Regarding the multimodal dataset, Figure~\ref{fig: avg_perf_mm_category} details the comparative performance between GPT-4V and Gemini Pro Vision across different question types, in alignment with the guidelines of the VCR dataset~\citep{zellers2019recognition}. We concentrate on the “Q $\rightarrow$ A” subtask as it most directly assesses the models' visual commonsense capabilities. Considering the data sample for each type, Gemini Pro Vision's performance either matches or is slightly lower than GPT-4V's, except in temporal-type questions, where it surpasses GPT-4V. This suggests its enhanced capability not only in recognizing but also in contextualizing time-related elements within visual scenarios.

\subsection{Reasoning Justification within MLLMs}
\begin{figure}[h]
\centering
\includegraphics[width=0.5\textwidth]{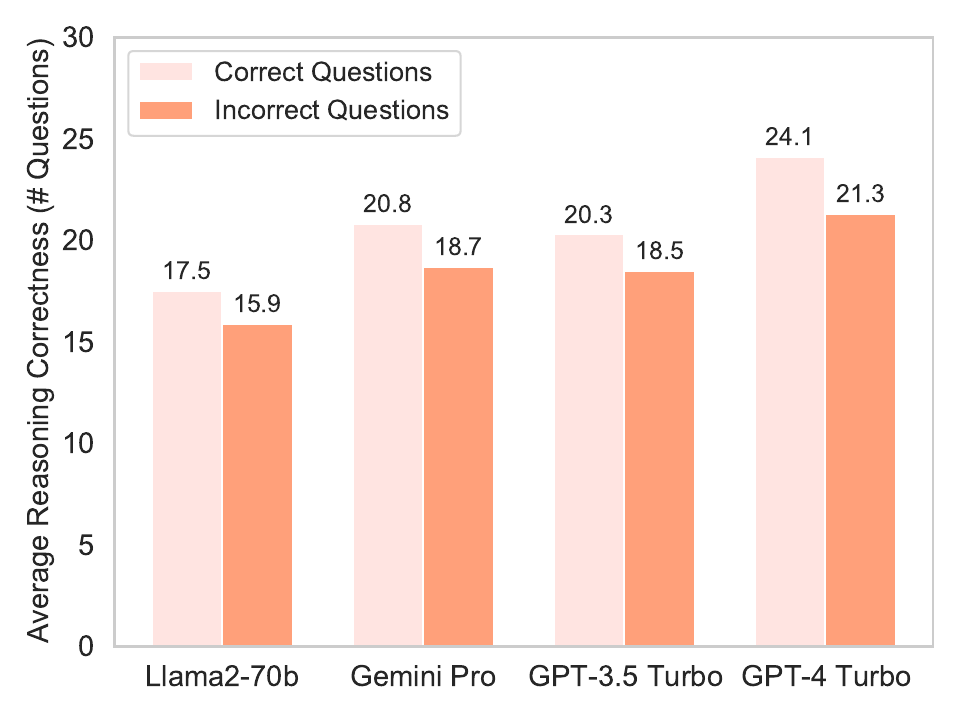}  
\caption{Average reasoning correctness across 11 language datasets. The comparison among four LLMs is based on a random sample of 30 correct and 30 incorrect questions per dataset. In cases where a dataset contained fewer than 30 incorrect questions, the data were scaled up to maintain consistency in the sample size.}
\label{fig: reasoning_correctness}
\end{figure}

\begin{figure}[htbp]
     \centering
     \begin{subfigure}[b]{0.48\textwidth}
         \centering
         \includegraphics[width=\textwidth]{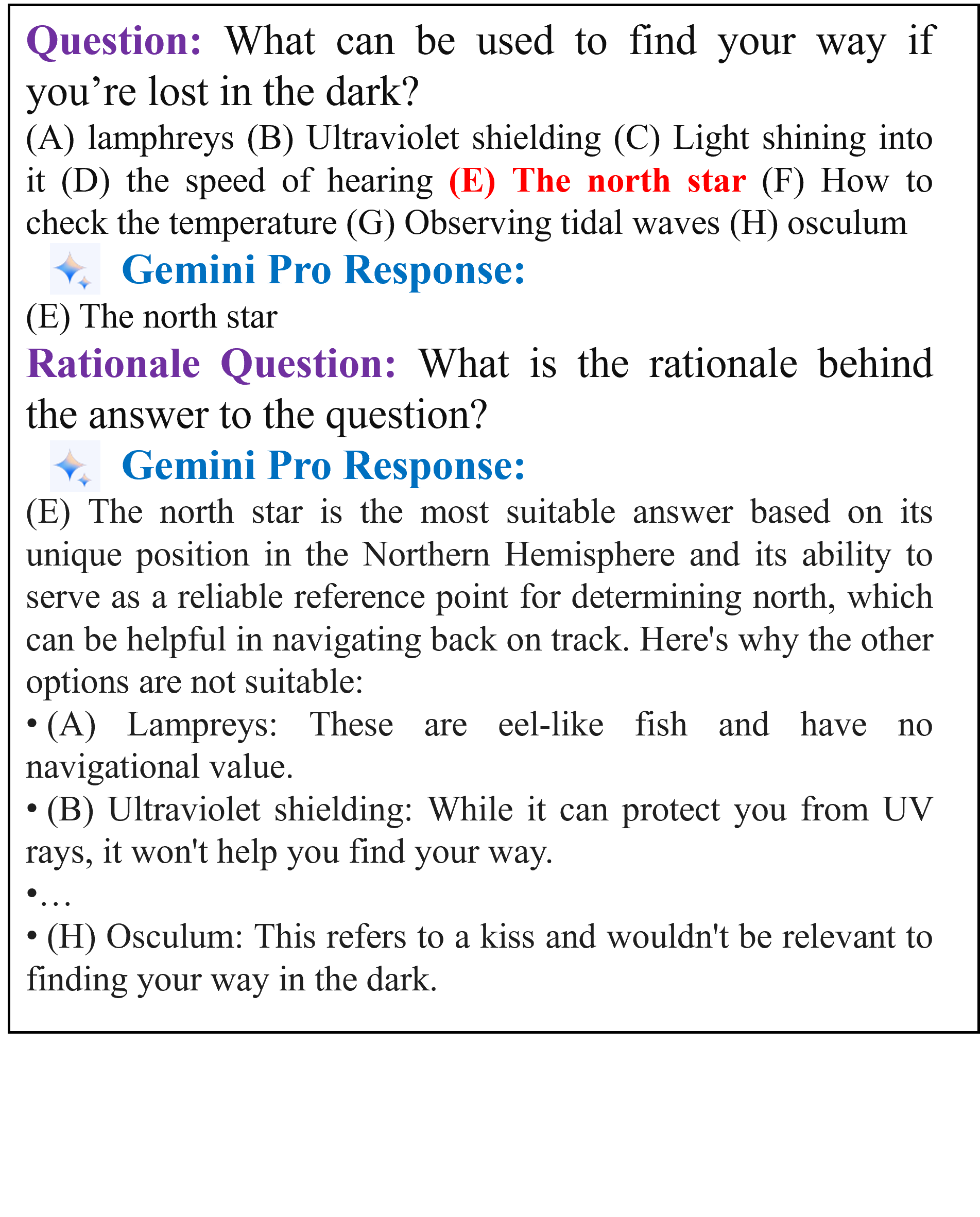}
         \vspace{-1.5cm}
         \caption{Example of a correct response and rationale explanation from Gemini Pro.}
     \end{subfigure}
     \hfill
     \begin{subfigure}[b]{0.48\textwidth}
         \centering
         \includegraphics[width=\textwidth]{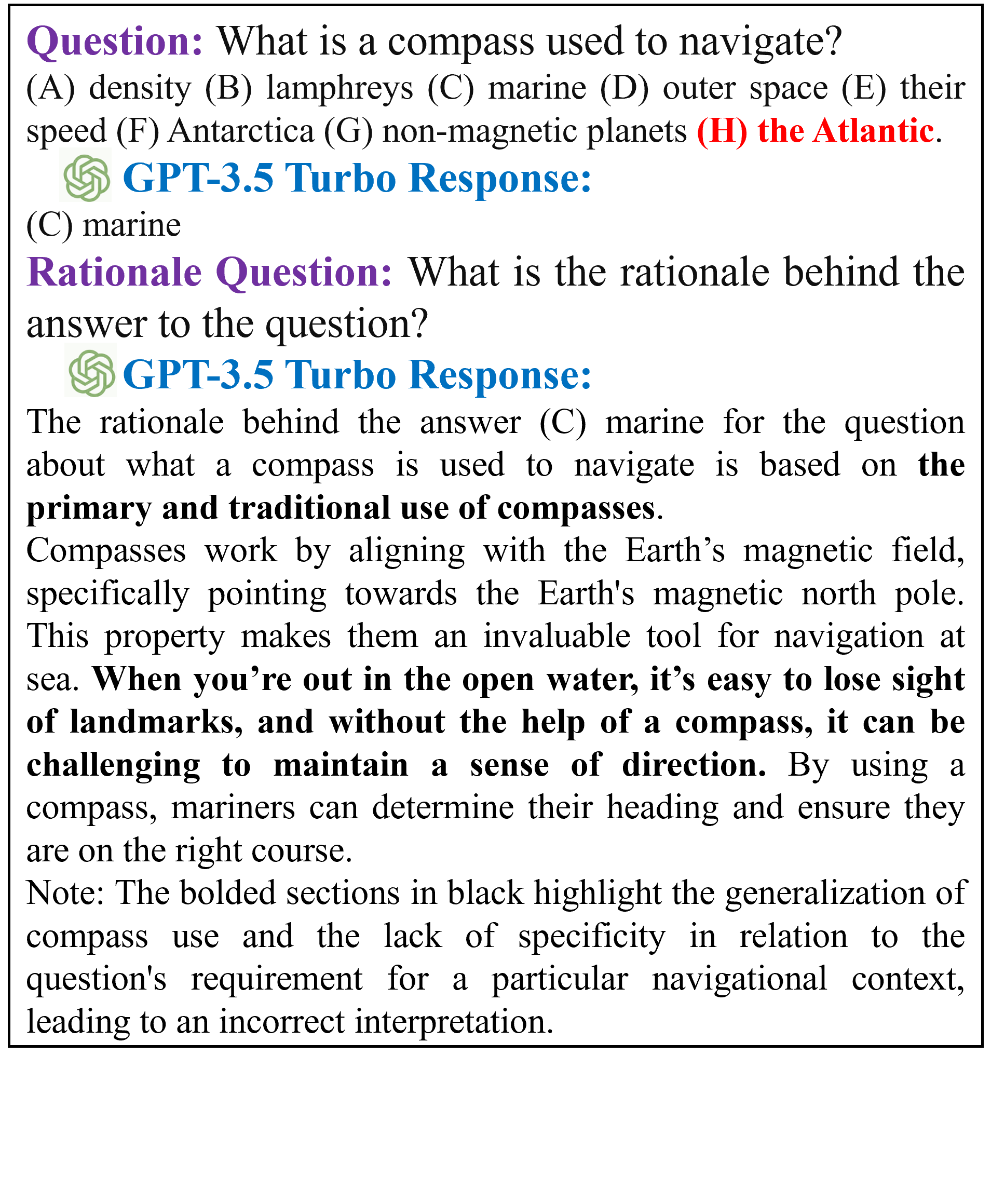}
         \vspace{-1.5cm}
         \caption{Example of an incorrect response and rationale explanation from GPT-3.5 Pro.}
     \end{subfigure}
        \caption{Model reasoning correctness justification examples. The sample questions are from the QASC dataset, with the correct answers highlighted in bold red. In example (a), Gemini Pro exhibits methodical reasoning by exclusion, carefully considering all options to reach the most logical conclusion. Conversely, example (b) illustrates GPT-3.5 Turbo's tendency towards unconventional logic, which can result in imaginative yet atypical answers. These instances emphasize the diverse strategies different models apply to commonsense reasoning tasks, revealing their distinct capabilities and limitations in such contexts.}
        \label{fig: correctness_examples}
\end{figure}

To assess the reasoning capabilities of MLLMs, particularly their ability to provide not only correct answers but also sound and contextually grounded reasoning in matters of commonsense, we adopted a systematic sampling approach. For each of the 11 language-based datasets evaluated with four LLMs, we randomly selected 30 questions that were correctly answered and 30 questions that were incorrectly answered by each LLM following~\citep{bian2023chatgpt}. In cases where a dataset presented fewer than 30 incorrect answers, we included all available incorrect responses to ensure comprehensive analysis. After selecting these questions, we prompted each model to explain “\textit{What is the rationale behind the answer to the question?}” The reasoning processes provided by the models were then manually reviewed and classified as either True or False, based on their logical soundness and relevance to the question. Figure~\ref{fig: reasoning_correctness} illustrates a comprehensive view of the average reasoning correctness across the 11 datasets, in terms of the sampled correct and incorrect questions. In fact, not every model had 30 incorrect questions for each dataset. In such scenarios, we scaled the available data up to 30 questions to ensure standardized computation. Figure~\ref{fig: reasoning_correctness} shows that GPT-4 Turbo's leading performance in both correct and incorrect answers highlights its advanced reasoning mechanisms and its ability to maintain coherent logic, even when the final answers are not accurate. Additionally, Gemini Pro has emerged as a notably proficient model, generally demonstrating commendable reasoning abilities and offering a well-rounded approach to commonsense reasoning. GPT-3.5, while trailing slightly behind Gemini Pro, still demonstrates competitive reasoning abilities. Figure~\ref{fig: correctness_examples} presents two real examples from Gemini Pro and GPT-3.5, illustrating the cases of a correct answer with a correct rationale and an incorrect answer with an incorrect rationale, respectively.

Moving to the multimodal perspective, our analysis of GPT-4V and Gemini Pro Vision on the VCR dataset reveals notable patterns in reasoning correctness. With GPT-4V at 24\% and Gemini Pro Vision at 26\%, approximately one-quarter of the cases showed both models correctly identifying the answers but failing to provide appropriate rationale. This discrepancy suggests that while the models can often determine the correct outcomes, their ability to understand or explain the underlying reasoning behind these answers is not consistently aligned. Furthermore, in the instances of incorrect answers, GPT-4V and Gemini Pro Vision showed correct rationales 16\% and 22\% of the time, respectively. This indicates that, despite arriving at incorrect conclusions, the models demonstrate a capacity for effective reasoning or logical processing. However, this does not consistently translate into accurate outcomes, implying that while some aspects of the required knowledge are captured, other crucial elements are likely missed. 

\subsection{Case Study: Gemini Pro in Commonsense}
Given our focus on evaluating the commonsense reasoning capabilities of the Gemini Pro model, we conduct a qualitative analysis to assess its performance across representative examples in four major categories (three language-based and one multimodal), as described in Section~\ref{data_section}. To ensure an authentic end-to-end capability evaluation, we present examples under the zero-shot learning setting, employing standard prompting techniques.

\noindent \textbf{General (CommonsenseQA).}
In the general commonsense evaluation (General and Contextual Reasoning category) using the CommonsenseQA dataset, consider the example question: “People are what when you're a stranger? (A) train (B) strange (C) human (D) stupid (E) dangerous.” Gemini Pro correctly chose (B) “strange,” and its reasoning process is notable. It recognized that while all options relate to the concept of a “stranger”, only “strange” accurately encapsulates the neutral and open-ended nature of the question. The model effectively ruled out other options: (A) “train”, for being too specific and unrelated; (C) “human”, as accurate but not capturing the question's essence; (D) “stupid”, for being judgmental and offensive; and (E) “dangerous”, due to its negative connotation. This selection of “strange” demonstrates an understanding of the unfamiliar nature associated with strangers, highlighting Gemini Pro's capability in interpreting and applying general commonsense knowledge appropriately.

\noindent \textbf{Temporal (TRAM).}
In the temporal commonsense evaluation (Specialized and Knowledge Reasoning category) using the TRAM dataset, consider the example question: “He also promises to ‘come to’ him. How long does it take for him to ‘come to’ him? (A) 100 years (B) in a minute's time (C) a few hours.” Lacking sufficient context, especially regarding the identities involved and the meaning of ‘come to’, Gemini Pro was unable to provide a definitive answer. Gemini Pro's response highlights a significant aspect of its temporal reasoning capabilities. It illustrates the model's reliance on specific contextual information to make accurate temporal judgments. While this cautious approach is prudent to avoid incorrect assumptions, it also signifies a limitation in addressing ambiguous or incomplete information – a frequent challenge in real-world communications. This example underlines the difficulties LLMs encounter in temporal reasoning tasks, especially when faced with scenarios that offer limited or unclear context.

\noindent \textbf{Social (Social IQa).}
In assessing Gemini Pro's performance in social commonsense reasoning using the Social IQa dataset (Social and Ethical Reasoning category), an interesting scenario was presented: “The people bullied Sasha all her life. But Sasha got revenge on the people. What will the people want to do next? (A) Do whatever Sasha says (B) Get even (C) Flee from Sasha.” The correct answer is (C), but Gemini Pro’s response is insightful. It chose (B) “Get even” as the most likely option, reasoning that the desire for revenge is a strong motivator and Sasha's actions likely ignited a similar desire in them. Gemini Pro considered (A) as a less likely option, depending on whether Sasha’s revenge instilled deep fear and assumed complete submission. The least likely option, according to Gemini Pro, was (C), which it associated with physical harm or an ongoing threat. This response demonstrates Gemini Pro's nuanced understanding of social dynamics and emotional motivations. However, it also highlights a limitation in accurately predicting human reactions in complex social scenarios, where emotional responses might not always follow a logical pattern. This instance reveals the challenges LLMs face in accurately interpreting and responding to intricate social situations, an area that remains challenging for AI systems.

\begin{figure}[h]
\centering
\includegraphics[width=0.5\textwidth]{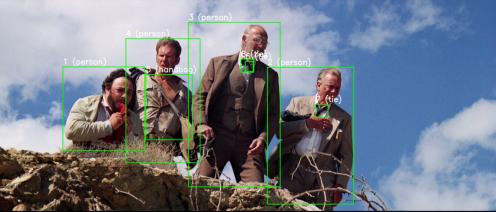}  
\caption{Example image from the VCR dataset.}
\label{fig: VCR_case_example}
\end{figure}

\noindent \textbf{Visual (VCR).}
In the visual commonsense evaluation using the VCR dataset, we analyzed Gemini Pro Vision's response to a scenario involving physical safety and potential danger, as shown in Figure~\ref{fig: VCR_case_example}. Presented with an image of individuals on the edge of a cliff, the model was questioned: “What would happen if person 4 pushed person 3 at this moment?” In this context, Gemini Pro Vision's response mirrored the logical inference that if the second person from the left (person 4) pushed the third person from the left (person 3), the result would be person 3 falling off the cliff, leading to a fatal outcome. This example from the VCR dataset underscores Gemini Pro Vision's ability to analyze visual scenes and make predictions about the potential consequences of actions within those scenes, a crucial aspect of visual commonsense reasoning. It demonstrates the model's grasp of spatial relations and physical consequences, providing evidence of its capacity to process and reason about complex visual information akin to human cognition.

Overall, the cases presented underscore the advanced reasoning capabilities of Gemini Pro and Gemini Pro Vision, while also identifying challenges in achieving human-like inference. These insights point to potential avenues for the continued enhancement of LLMs and MLLMs.

\subsection{Error Analysis} 

To gain a deeper understanding of the mistakes made by models, we manually analyzed instances where a model made incorrect choices or provided inappropriate answers. We conducted a thorough examination of common error types encountered in commonsense reasoning tasks, with the results averaged across four LLMs. Our focus was on assessing these models in two distinct settings: zero-shot SP and few-shot CoT. Table~\ref{tab:error_types} presents the proportions of five common error types observed in each setting, with the data averaged over the four LLMs.

Context misinterpretation emerged as the most frequent error, occurring more often in the zero-shot SP setting (28.6\%) compared to the few-shot CoT (23.4\%). This trend suggests that the additional context in few-shot CoT helps models better understand scenarios, thereby reducing errors related to contextual misunderstanding. Logical errors were the second most common, accounting for 23.9\% in zero-shot SP and slightly less in few-shot CoT (20.1\%), indicating that extra examples in the latter setting aid in more consistent logical reasoning. Ambiguity errors, at 16.2\% in zero-shot SP, were reduced to 11.6\% in few-shot CoT, demonstrating the effectiveness of added context in resolving language ambiguities. In contrast, Overgeneralization errors showed an increase in few-shot CoT (15.6\%) from zero-shot SP (11.8\%), possibly due to models' overextending patterns learned from the additional examples. Notably, knowledge errors, where models misapplied correct and necessary commonsense knowledge, saw a significant increase in few-shot CoT (29.3\%) compared to zero-shot SP (19.5\%). This finding suggests that while extra context can be beneficial, it can also lead to inaccuracies, particularly in complex or nuanced scenarios.

\begin{table}[htbp]
\centering
\caption{Proportion of common error types in commonsense reasoning in LLM evaluation. Misinterpret. represents misinterpretation.}
\label{tab:error_types}
\resizebox{\linewidth}{!}{%
\begin{tabular}{l c c}
\hline
\textbf{Error Type} & \textbf{Zero-shot SP} & \textbf{Few-shot CoT}  \\
\hline
Context Misinterpret. & 28.6\% & 23.4\% \\
Logical Errors & 23.9\% & 20.1\% \\
Text Ambiguity & 16.2\% & 11.6\% \\
Overgeneralization & 11.8\% & 15.6\% \\
Knowledge Errors  & 19.5\% & 29.3\% \\
\hline
\end{tabular}}
\end{table}

In our analysis of the VCR dataset, we focused on instances where either GPT-4V or Gemini Pro Vision chose incorrect answers in the Q $\rightarrow$ A subtask. The four common error types for each model are summarized in Table~\ref{tab:visual_error_types}. Emotion recognition errors were the most common, with GPT-4V encountering these errors in 30.1\% of cases and Gemini Pro Vision slightly more at 31.3\%. This high incidence suggests that both models find interpreting emotional cues in visual content particularly challenging, underscoring the complexity of deciphering human emotions from visual stimuli. Spatial perception errors were also significant, constituting 22.5\% of errors for GPT-4V and 25.2\% for Gemini Pro Vision. These figures indicate the models' difficulties in accurately understanding spatial relationships and the arrangement of elements in images. Logical errors were another major error type, more pronounced in GPT-4V (27.7\%) than in Gemini Pro Vision (24.9\%), pointing to challenges in logical reasoning within visual contexts. Context misinterpretation, although less frequent, was still a notable issue, with GPT-4V at 19.7\% and Gemini Pro Vision at 18.6\%. These errors demonstrate the models' struggles with grasping the overarching context or narrative depicted in the visual content.

Overall, error analysis  sheds light on the specific challenges LLMs and MLLMs face in commonsense reasoning, providing valuable insights for future improvements for future model refinement.

\begin{table}[htbp]
\centering
\caption{Proportion of commmon error types in visual commonsense reasoning in MLLM evaluation (GPT-4V and Gemini Pro Vision). Misinterpret.: and E. represent misinterpretation and errors, respectively.}
\label{tab:visual_error_types}
\resizebox{\linewidth}{!}{%
\begin{tabular}{l c c}
\hline
\textbf{Error Type} & \textbf{GPT-4V} & \textbf{Gemini Pro Vision}  \\
\hline
Context Misinterpret. & 19.7\% & 18.6\% \\
Spatial Perception E. & 22.5\% & 25.2\% \\
Emotion Recognition E. & 30.1\% & 31.3\% \\
Logical Errors & 27.7\% & 24.9\% \\
\hline
\end{tabular}}
\end{table}

\section{Related Work}

\noindent \textbf{Commonsense Reasoning in NLP.}
Commonsense reasoning has gained renewed attention in recent years, especially in the context of advancements in LLMs that have significantly influenced numerous applications in NLP. However, there is a growing concern about their ability to understand and reason about commonsense knowledge~\citep{storks2019commonsense, tamborrino2020pre, bhargava2022commonsense}. This concern is echoed in various studies that focus on evaluating the commonsense reasoning capabilities of LLMs~\citep{bian2023chatgpt, weng2023large, shen2023experimental}. Concurrently, researchers have been exploring diverse strategies to enhance the commonsense reasoning capabilities of NLP systems. These strategies range from leveraging large-scale knowledge graphs to employing methods of commonsense knowledge transfer, aiming to endow NLP systems with a deeper and more nuanced understanding of commonsense concepts~\citep{huang2023mvp, ye2023improving, zhou2023commonsense}. Prior to delving into methodological refinements, a comprehensive evaluation is essential to understand the authentic commonsense reasoning capabilities of LLMs. In our study, we endeavor to advance this line of inquiry by examining how LLMs, particularly focusing on the Gemini model, navigate and implement commonsense reasoning in various NLP contexts.

\noindent \textbf{Training Paradigms in LLMs.} In NLP research, pretraining language models on large-scale varied textual datasets has become essential. This approach endows models with a comprehensive knowledge base across numerous fields. Initially, leveraging this knowledge often involved fine-tuning models with task-specific data. BERT-based models like BERT~\citep{kenton2019bert} and RoBERTa~\citep{liu2019roberta} exemplify this, being applied to tasks ranging from disease prediction~\citep{zhao2021empirical} to text classification~\citep{wang2022integrating} and time series analysis~\citep{wang2022enhancing}. The debut of GPT-3 (Brown et al., 2020) shifted this focus towards more flexible learning methods like zero-shot and few-shot learning, showcasing models' adaptability to new tasks with minimal data~\citep{brown2020language}. This shift has spurred the development of novel prompting techniques to enhance LLMs' reasoning and understanding capabilities, including chain-of-thought (CoT) prompting~\citep{wei2022chain}, self-consistency with CoT~\citep{wang2022self}, tree-of-thought prompting~\citep{yao2023tree}, and metacognitive prompting~\citep{wang2023metacognitive}. In this work, we establish evaluations by considering four popular LLMs for language-based tasks under zero-shot and few-shot settings, and two MLLMs for multimodal tasks under the zero-shot learning paradigm. Our goal is to provide an in-depth understanding of their strengths and limitations in diverse commonsense reasoning tasks.

\noindent \textbf{Evaluations on MLLMs.}
Since the release of the state-of-the-art MLLM, GPT-4V, several evaluations have been conducted across diverse tasks, including medical imaging~\citep{wu2023can}, visual question answering ~\citep{li2023comprehensive, yang2023dawn}, and video understanding~\citep{lin2023mm}. These evaluations typically focus either on case-by-case qualitative analyses through example demonstrations or on quantitative assessments by analyzing the model's performance across diverse tasks. The recent release of Google's Gemini has garnered considerable attention, and early experiments have been conducted to evaluate its capabilities in both language understanding~\citep{akter2023depth} and the multimodal domain~\citep{liu2023evaluation, fu2023challenger}. However, a significant gap remains in fully comprehending the commonsense reasoning capabilities of Gemini, a known potential shortcoming since its introduction. In our work, we conduct a comprehensive analysis of Gemini's capabilities in this area, along with comparisons to other leading MLLMs, thereby highlighting both the potential and areas for further improvement in future research.

\section{Discussion}
In this study, we conducted a comprehensive evaluation of state-of-the-art LLMs and MLLMs, focusing particularly on Gemini Pro and Gemini Pro Vision, across 12 diverse commonsense reasoning datasets. Our findings indicate that while these models mark a significant advancement in various domains, demonstrating impressive performance in commonsense reasoning tasks, they still exhibit limitations, particularly in tasks requiring deep contextual understanding or abstract reasoning, such as those involving temporal dynamics, riddles, or intricate social scenarios.

Looking ahead, addressing these challenges is crucial to enhance the overall effectiveness of LLMs and MLLMs in commonsense reasoning. Future research should aim to refine the models' capabilities in interpreting and reasoning within complex contexts and abstract scenarios. Additionally, there is an emerging need for more holistic evaluation metrics and methodologies capable of accurately assessing the nuances of commonsense reasoning in AI systems. These metrics should evaluate not only the correctness of responses but also their logical coherence and context relevance.

In conclusion, our study underscores that perfecting commonsense reasoning in LLMs and MLLMs remains an ongoing endeavor. The observed performance discrepancies among these models reveal intriguing areas for further research and improvement. Although significant progress has been made, achieving AGI still represents a substantial goal on the horizon. Our work lays the groundwork for future exploration in this field, highlighting both the achievements and the areas in need of enhancement in the realm of commonsense reasoning within LLMs and MLLMs.

\section{Limitations}
While this study offers valuable insights into the role of LLMs and MLLMs in commonsense reasoning, there are some limitations to our work. Firstly, our evaluation is heavily dependent on the selected questions and datasets used for analysis. Despite their diversity, these datasets may not cover all facets of commonsense reasoning. As a result, the performance and capabilities of Gemini Pro and other models could vary in real-world scenarios or with alternative datasets. Additionally, our analysis is confined to English language datasets, limiting the generalizability of our findings to other languages or multilingual contexts, where cultural nuances and linguistic differences are crucial in commonsense reasoning. Finally, our study represents a specific moment in the rapidly evolving landscape of AI, focusing on API-based systems that are subject to change. The introduction of newer models or updates to existing ones might lead to different performance outcomes, highlighting the need for ongoing evaluation and analysis.

\bibliography{anthology,custom}
\bibliographystyle{acl_natbib}

\appendix

\end{document}